\documentclass[conference]{IEEEtran}
\IEEEoverridecommandlockouts
\usepackage{cite}

\usepackage{booktabs}

\usepackage{amsmath,amssymb,amsfonts}
\usepackage{algorithmic}
\usepackage{graphicx}
\usepackage{textcomp}
\usepackage{xcolor}
\def\BibTeX{{\rm B\kern-.05em{\sc i\kern-.025em b}\kern-.08em
    T\kern-.1667em\lower.7ex\hbox{E}\kern-.125emX}}
\begin{document}

\title{Efficient Object-centric Representation Learning with Pre-trained Geometric Prior}

\author{\IEEEauthorblockN{Phúc H. Le Khac, Graham Healy and Alan F. Smeaton}
\IEEEauthorblockA{Dublin City University, Glasnevin, Dublin 9, Ireland\\
Email: alan.smeaton@dcu.ie}}


\maketitle

\begin{abstract}
This paper addresses key challenges in object-centric representation learning of video. While existing approaches struggle with complex scenes, we propose a novel weakly-supervised framework that emphasises geometric understanding and leverages pre-trained vision models to enhance object discovery. Our method introduces an efficient slot decoder specifically designed for object-centric learning, enabling effective representation of multi-object scenes without requiring explicit depth information. Results on synthetic video benchmarks with increasing complexity in terms of objects and their movement, object occlusion and camera motion demonstrate that our approach achieves comparable performance to supervised methods while maintaining computational efficiency. This advances the field towards more practical applications in complex real-world scenarios.
\end{abstract}


\section{Introduction}
\label{sec:intro}

Object-centric learning in video aims to learn a structured representation of a visual scene, by decomposing the input signal into its natural composition of \textit{objects} \cite{greff2020BindingProblemArtificial}. This allows for multiple or multimodal representations of a visual scene, from descriptor tags or captions, to details of scene composition including objects and their relationships.
An object-centric learning method should be designed in a way that results in a set of object representations for visual content that are independent yet compatible with each other, while together completely representing the input scene.
Such object-centric representations represent a promising direction to improve tasks from visual scene understanding \cite{chen2017LearningObjectcentricTransformation} to planning and reasoning \cite{ding2021AttentionLearnedObject}.

Object-centric representation learning has progressed significantly in recent years, evolving from proof-of-concept methods in synthetic 2D scenarios with simple geometric shapes \cite{kipf2019ContrastiveLearningStructured} to handling more intricately rendered 3D environments \cite{locatello2020ObjectCentricLearningSlot}.
However, it faces challenges when applied to more realistic and complex datasets \cite{kipf2021ConditionalObjectCentricLearning}.
In particular, when using image autoencoding as the pre-training task, object-centric models tend to rely heavily on low-level RGB colour values for both the reconstruction and object-discovery tasks, limiting their scalability to  complex visual scenes.
In addition, scaling up the current object-centric representation methods is also  computationally expensive.

Decoders for object-centric learning  can be categorised into two groups, slot-based independent decoding or set-based joint decoding, each with advantages and disadvantages. Some promote independence and compatibility of slots but are less powerful and require more compute time and memory during training while  others are more powerful but lack the inductive bias that is helpful for learning object-centric representations.

In recent years, advances in learning visual representations and deep learning  have  been driven by scalable architectures like the Vision Transformer (ViT) \cite{dosovitskiy2020ViT}, efficient hardware utilisation \cite{he2022MaskedAutoencodersAre}, and the availability of large and diverse datasets \cite{singh2023EffectivenessMAEPrepretraining}.
Motivated by these advances,  in this paper we decouple the challenges of learning \textit{object} representations from that of learning \textit{visual} representations, capitalising on the  progress of each sub-field.
In particular, our attention is directed towards a specific category of visual representation that incorporates a richer set of geometric information.

The  contribution of this paper lies in our   demonstration of effective object-centric learning techniques.
In addition, we propose a simple extension of an object decoder which uses an attention mechanism to allow rich interaction between slots in the latent space, while the more expensive visual decoder only needs one iteration to reconstruct  the high-dimensional input.


\section{Related Work}

Our work belongs to the broader field of visual representation learning which involves training neural networks on visual data to obtain useful and compact vector-based representations that are applicable to various tasks and complementary to other representations such as descriptor tags or captions.
Contrastive learning \cite{le-khac2020ContrastiveRepresentationLearning} is a promising method for visual representation learning, wherein features emerge from the comparison between similar and dissimilar input pairs.
Recent efforts aim to alleviate the need for negative samples and the large batch sizes required in contrastive methods \cite{grill2020BootstrapYourOwn}. For instance, DINO \cite{caron2021DINO} employs a discriminative self-supervised pre-training approach via self-distillation, akin to contrastive learning but without explicit negative pairs. DINO has exhibited intriguing emergent properties, such as object-segmentation masks, arising from discriminative pre-training when coupled with ``global-to-local" correspondence.
In addition to discriminative or predictive methods, techniques based on the reconstruction objective, such as Masked Autoencoders (MAE) \cite{he2022MaskedAutoencodersAre}, serve as robust visual representation learners.
All such MAE-based methods \cite{girdhar2023OmnimaeSingleModel, weinzaepfel2022CroCoSelfsupervisedPretraining} learn global visual representations by randomly masking portions of the input and learning to reconstruct the masked regions.

Object-centric representation learning revolves around the fundamental binding problem \cite{greff2020BindingProblemArtificial}, that requires a learning system to simultaneously discover and bind information about independent objects into separate representation slots.
Earlier approaches such as \cite{greff2017NeuralExpectationMaximization} treat the problem of detecting object instances as a form of perceptual grouping and clustering.
Similar to our work, extending object-centric methods to dynamic visual environments in videos is an important and popular line of research with noteworthy approaches such as  SIMONe which \cite{kabra2021SIMONeViewinvariantTemporallyabstracted} decomposes a multi-object video into object  and frame latents that can be composed from different video sources.

Scaling object-centric methods to larger datasets has proven to be a formidable challenge.
While the approach of autoencoding visual inputs works effectively on visually simple datasets with a limited range of colour variations, it struggles when applied to more realistic datasets.
The work most closely related to ours is SAVi \cite{kipf2021ConditionalObjectCentricLearning} and SAVi++ \cite{elsayed2022SAViEndtoendObjectcentric}, which uses the efficient and scalable Slot Attention \cite{locatello2020ObjectCentricLearningSlot} to learn a set of object latents initialised from that of  previous frames which were initially  conditioned with extra signals such as bounding boxes or object masks.
This work also pioneered the uses of extra modalities such as optical flow and depth information as a training signal to overcome the limitation of pixel-based reconstruction training.
DINOSAUR \cite{seitzer2023BridgingGapRealWorld} is a recent work which attempts to discover and learn object-centric representations using the semantic representation obtained from the self-supervised visual representation of DINO \cite{caron2021DINO}.

While some methods for object-centric video processing are based on contrastive learning \cite{le-khac2020ContrastiveRepresentationLearning}, the majority of architectures employ the autoencoder framework which makes the decoder all the more important.
The most popular approach known as ``slot-wise decoder'' or ``mixture of component decoder''~\cite{kabra2021SIMONeViewinvariantTemporallyabstracted}  reconstructs the final output by weighting multiple reconstructions, each derived from a single slot within the set of object representations.
Consequently, for one training sample, the decoder needs to process more forward and backward passes proportional to the number of slots, requiring significantly more compute time and memory. 

While the slot-based decoder encourages independence and compatibility between slots since they are decoded in parallel, the limited interaction of slots at the output level can be restrictive for the reconstruction objective.
Consequently, others 
have explored deviating from the slot-based decoding approach and instead use a standard efficient Transformer 
to decode from the entire set representation.

\section{Method}


\subsection{Pre-trained visual encoder with geometric representation} \label{sec:geometric}

To scale up the process of learning object-centric representations to diverse and visually complex datasets, we  leverage the visual representation from a pre-trained model as the backbone to bootstrap the learning.
Given the proliferation of pre-trained vision models in recent years, the question arises: which pre-trained model should be employed for object-centric learning, and what motivates this choice?

\paragraph{Emergence of objects from semantics}
Most closely related to our approach is DINOSAUR \cite{seitzer2023BridgingGapRealWorld}, which learns object-centric representations from the semantic representation obtained from the self-supervised pre-trained model DINO \cite{caron2021DINO}.
DINO is a  powerful vision model that is effective for various downstream tasks.
Notably, it demonstrated for the first time the notion of an \emph{object} emerging from self-supervised training without any explicit label, as revealed by inspecting its attention map in the intermediate layer.
These representations were hypothesised  to be especially useful for the object discovery task in object-centric learning.

\paragraph{Emergence of objects from geometry}

The human and other animal's binocular vision systems introduce a strong bias to represent our world in three dimensions where we can perceive a hierarchy of representations with independent objects which can cover or mask each other, or  can collide and alter trajectory.
Geometric features have also been known to be helpful for many visual tasks \cite{wang2023DepthaidedCamouflagedObject}.
We hypothesise that a pre-trained representation with rich \emph{geometric information} of an underlying 3D scene in addition to \emph{semantic information}, would be beneficial for the goal of object-centric learning.

CroCo \cite{weinzaepfel2022CroCoSelfsupervisedPretraining} is a pre-trained vision model based on the framework of Masked Autoencoding (MAE) \cite{he2022MaskedAutoencodersAre}.
CroCo extends the MAE masked image patches to binocular settings, where the context from a different view of the same scene is used to reconstruct the masked patches from a different view.
This cross-view completion mechanism endows a network with the ability to process stereo input, and similar to the development of human cognition,  leads to a representation space with strong geometric information.

\begin{figure}[htb]
    \begin{center}
        \includegraphics[width=\linewidth]{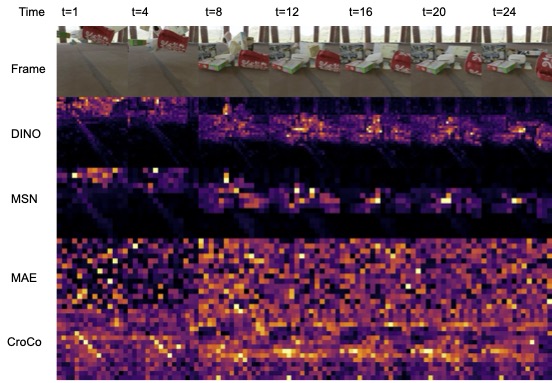}
    \end{center}
    \caption{Attention map of different pre-trained self-supervised vision models on a video from the Movi-C dataset. While DINO and MSN show localised attention towards foreground objects, MAE and CroCo exhibit a diffused, global attention map.}\label{fig:attn-map}
\end{figure}

 \paragraph{Attention Maps Mostly Indicate Semantic Representations}
An initial step to validate whether the representation of CroCo \cite{weinzaepfel2022CroCoSelfsupervisedPretraining} is suitable for object-centric learning would be to check whether the same notion of object is captured in its attention patterns.
In Fig.~\ref{fig:attn-map}, we visualise the attention maps from the last layer of the encoder, generated using various pre-trained visual models across different frames of a multi-object video sourced from the MOVi-C dataset \cite{greff2022KubricScalableDataset}.
Our analysis of four distinct pre-trained models reveals discernible characteristics which we  classify into two distinct groups.
The first group comprises DINO \cite{caron2021DINO} and MSN \cite{assran2022MaskedSiameseNetworks}, both of which exhibit a pronounced focus on the objects within the scene. Given our scene's composition with numerous small objects, the attention maps of DINO and MSN collectively concentrate on what could be described as the foreground.

The second group, including MAE \cite{he2022MaskedAutoencodersAre} and CroCo \cite{weinzaepfel2022CroCoSelfsupervisedPretraining}, do not display any conspicuous grouping behaviour as their attention scores disperse across the feature maps, extending even to areas corresponding to background elements in the original input images.
Despite this apparent ``bug" in their attention maps, these models demonstrate strong performance after fine-tuning, remaining competitive with models from the first group.

One possible explanation for this divergence in attention style lies in their pre-training paradigms.
While MSN and DINO employ self-supervised discriminative methods, MAE and CroCo rely on unsupervised generative models.
The first group can trace their lineage back to earlier contrastive methods, albeit without the use of negative pairs, while the second group adopts an autoencoding framework, incorporating input-masking techniques.

Since the seminal work of DINO \cite{caron2021DINO}, which demonstrated the emergence of objects in the attention maps of self-supervised training models, numerous subsequent works have capitalised on this insight to incorporate ``objectness" into their own models. 
This  might lead one to assume that for object discovery and learning, methods like DINO or MSN should be preferable.
Counterintuitively however, our experiments demonstrate that this is not the case. 

 \subsection{Attentional Slot Decoder}\label{sec:decoder}

 \paragraph{Slot-based vs. Set-based Object Decoder:}
In the literature, there are two main types of decoder used for object-centric learning.
The first group is \emph{slot-based} or \emph{mixture of components} decoders such as the Spatial Broadcast Decoder. 
This  promotes compatibility among slots but  can be computationally expensive while being more restrictive and less powerful due to the limited interactions between the slot latents.
On the other hand, \emph{set-based} decoders like the Transformer Decoder 
can be more efficient to train, allowing for richer interactions between latent vectors, but does not enforce object-independence among slots.

\begin{itemize}
    \item Denote the set of $K$ slots as $ \mathbf S=[\mathbf s_1, \mathbf s_2, ..., \mathbf s_K] \in \mathbb R^{k \times d} $ obtained from the output of the slot encoder where each captures an independent part of the scene.

    \item Denote $\mathbf g \in \mathbb R^d$ as the optional \textit{global} scene representation. This global representation of the scene can be obtained by pooling from the grid of visual representations $\mathbf V \in \mathbb R^{h \times w \times d}$ or extracting from the special CLS token of the Transformer architecture.

    \item Decoder's queries:
        First we initialise a 2D spatial grid (3D spatial-temporal if working with video) with Fourier positional embeddings \cite{tancik2020FourierFeaturesLet}.
        At this stage, the decoder queries consist only of positional information of the pixels (or patches) that it corresponds to. This is the same for all input samples.

        To inject scene-specific information, we broadcast the global scene embedding $\mathbf g$ and add to every spatial dimension of $\mathbf Q$.

    \item Keys and values in attention mechanism:
        These are obtained from the set of object latents by using a linear layer followed by LayerNorm 
        $ \mathbf{K} = \mathrm{LN}(\mathbf{S} \mathbf{W}_K), \mathbf{V} = \mathrm{LN}(\mathbf{S} \mathbf{W}_V). $
        with $\mathbf W_K, \mathbf W_V \in \mathbb R^{d \times d}$ as the corresponding weight matrices for the linear layers.

    \item Cross-attention operation:
        We now perform a standard cross-attention operation with the set of queries, keys and values obtained.
        The softmax normalisation operation is performed over the key dimensions of the attention matrix $\mathbf A \in \mathbb R^{hw \times k} $.
        The final output of the cross-attention module is the weighted average of $\mathbf V$ based on the attention scores $\mathbf A$.

\end{itemize}

\noindent 
With a spatial or a spatial-temporal grid of embeddings, we  pass these through a visual decoder module to obtain the final reconstruction in the output space.
We follow standard practice and use a $1\times 1$ convolutional network, effectively a position-wise MLP, to decode these embeddings into RGB values $\mathbf O \in \mathbb R^{hw \times 3}$.
Crucially, we add a residual connection from the positional embedding  to the output of this attention mechanism.
We find it necessary to inject  positional information back into the decoder's embeddings in order for the model to be able to reconstruct and place objects in the correct location in the scenes.

The key insight into our approach is that we do not  measure the contributions of object embeddings in the output space by decoding an alpha mask. 
The key-normalised attention scores now serve as the alpha masks, which state which object  each pixel belongs to.



\subsection{Architecture}\label{sec:arch}
\begin{figure}
\centering 
  \includegraphics[width=\linewidth]{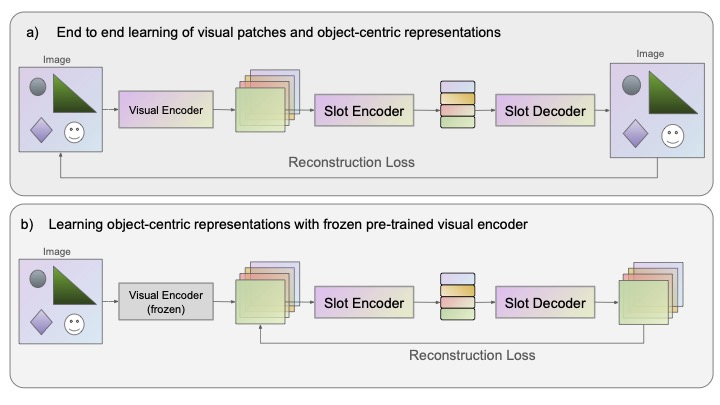}
  \caption{Our overall architecture and training pipeline including the reconstruction objective on the pre-trained geometric features and using an Attentional Slot Decoder for efficient learning.}
  \label{fig:arch}
\end{figure}

Our overall architecture for efficient learning of object-centric representations is shown in Fig.~\ref{fig:arch}, with a pre-trained visual decoder as the backbone and an attentional slot decoder.

Input images or videos $\mathbf X \in \mathbb R^{h\times w \times 3}$ are first encoded via a visual backbone.
We use the CroCo~\cite{weinzaepfel2022CroCoSelfsupervisedPretraining} pre-trained method since its representations contain more geometric information about the 3D world.
These visual embeddings $\mathbf V = \mathrm{ViT}(\mathbf X) \in \mathbb R^{h'\times w'\times d}$ are  downscaled by a factor of patch size $p$ from the chosen ViT \cite{dosovitskiy2020ViT}, $h' = h / p, w' = w / p$.

Following previous works, we use a Slot Attention \cite{locatello2020ObjectCentricLearningSlot} module to further obtain a set of $k$ object representations from the visual feature map: $\mathbf O = \mathrm{SlotAttention}(\mathbf V, \mathbf S) \in \mathbb R^{k\times d}$.
We use the Attentional Slot Decoder introduced in Section~\ref{sec:decoder} to decode the slot representations back to the encoded features of the backbone: $\mathbf{Y} = \mathrm{AttentionalSlotDecoder}(\mathbf{x}) \in \mathbb R^{h\times w\times d}$.
We compute the Mean Squared Error reconstruction loss on the \emph{feature space} of our pre-trained visual encoder: $\mathbb L = \| \mathbf Y - \mathbf V \|^2_2$.
If the pre-trained visual backbone has a corresponding pixel decoder, we can reconstruct the original input from the predicted representations $\hat{X} = \mathrm{Decoder}(\mathbf Y)$ for visualisation purposes.

\section{Experiments}


In our evaluation we focus on the most challenging and important aspects of object-centric learning, unsupervised discovery, representation and object tracking with minimal supervisory signals. Similar to SAVi~\cite{kipf2021ConditionalObjectCentricLearning}, we use the objects' bounding box information as the conditional information to initialise the set of slot latents.
At every decoded location we assess the influence of each object slot on the output using its corresponding attention scores. To establish the ground truth for prediction segmentation, we employ the \textit{argmax} operator across slots, i.e we take the index of the slot with the highest attention logit at each decoded location.

We use the Adjusted Rand Index (ARI) as the  metric for assessing the efficacy of video decomposition, object segmentation, and tracking.  ARI serves as a measure of clustering similarity, gauging the congruence between predicted segmentation masks and ground-truth masks in a manner that remains unaffected by permutations.
Similar to prior works \cite{greff2019MultiObjectRepresentationLearning, locatello2020ObjectCentricLearningSlot}, 
we compute the ARI for foreground objects, referred to as ARI-FG. In the context of video data, a singular cluster in the ARI calculation corresponds to the segmentation of an individual object over the entire video duration. This requires temporal coherence, with the absence of alterations in object identity, for achieving favourable outcomes on this metric.

\paragraph{Training details:}
We use multiple variants of the MOVi dataset \cite{greff2022KubricScalableDataset} consisting of synthetic multimodal videos  created specifically for the study and development of unsupervised multi-object video understanding.
MOVi consists of 5 variants, A to E, with increasing  complexity with more objects, more realistic backgrounds and more motion from objects and from the camera.
Being simulated and rendered programmatically, MOVi provides not only high-quality and realistic videos but also  rich and dense multimodality annotations for segmentation masks, depth, optical flow, surface normals and object coordinates. 

During the training phase, we cut 24-frame videos from the MOVi dataset into 4 consecutive clips of 6 frames each and treat the clips as independent samples in a training batch.
For each sample, we conditionally initialised the object slots with the object's bounding box of the first frame, and recurrently carry over the slots from the previous frame as the initial signal for the next.
Each of the 5 MOVi  variants contain a total of 10,000 videos, with 2,500 videos reserved for validation and  the remaining 7,500  used for training.
On the MOVi-A and MOVi-C datasets, we use 11 slots for the object latent representations, while for the MOVi-E variant we initialised the object latents with 24 slots.
We train on videos with a resolution of 128x128 for 50,000 steps on a single GPU with 24GB memory, utilising a batch size of 16.
The model is optimised using the Adam optimiser with an initial learning rate set at 0.0002.

\subsection{Attentional Slot Decoder} \label{sec:exp-dec}
To evaluate our architectural design for the slot decoder, we  follow the architecture of SAVi \cite{kipf2021ConditionalObjectCentricLearning}.

\begin{figure}[!htb]
    \begin{center}
        \includegraphics[width=\linewidth]{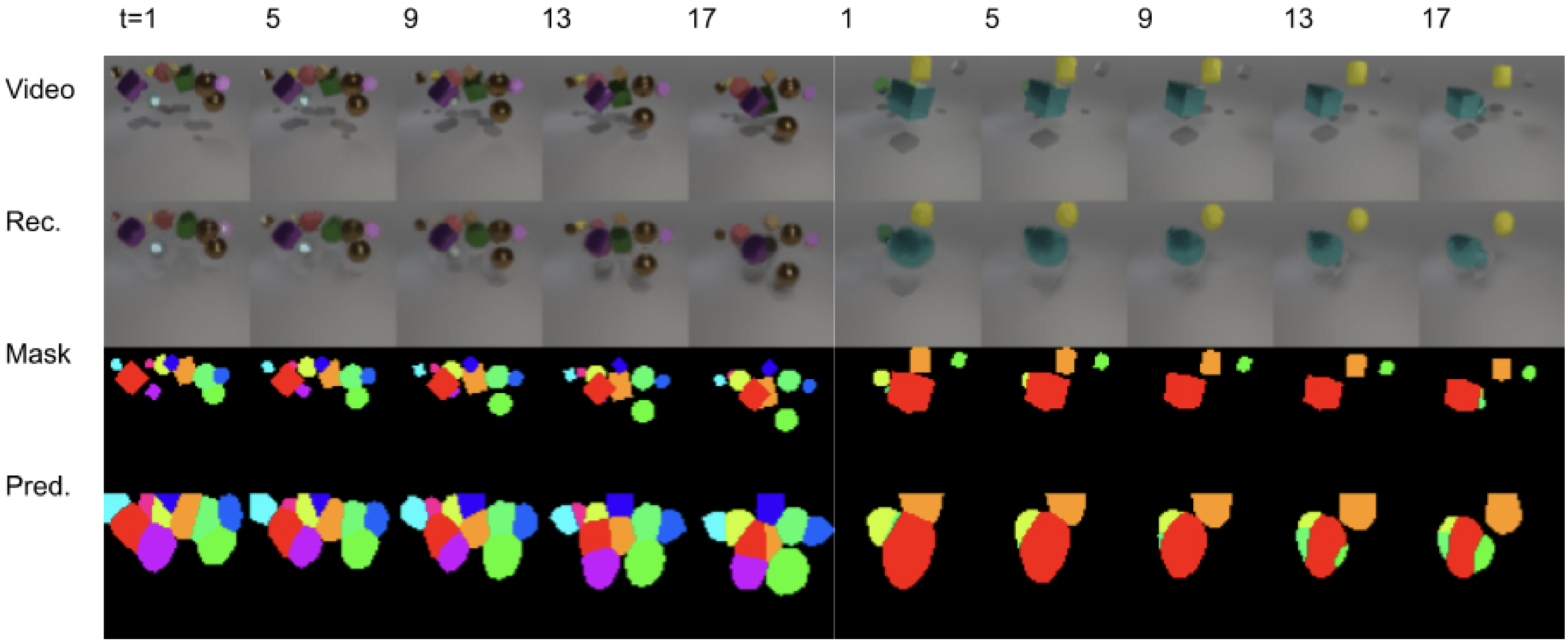}
    \end{center}
    \caption{Indicative qualitative results of our Attentional Decoder on the Movi-A dataset. In each row of images we visualise the input video, the RGB reconstruction target of the input (Rec.), the ground truth object masks (Mask) and the predicted object masks from our object representations (Pred.). 
    }
    \label{fig:dec-a-rgb}
\end{figure}
In Fig.~\ref{fig:dec-a-rgb} we visualise some samples from the MOVi-A dataset \cite{greff2022KubricScalableDataset} to show our  decoder is capable of decoding to RGB pixel values.
The  samples on the left are for a scene with many small objects with complicated interactions, yet our method can still reconstruct the original video and is able to segment most of the objects based on the slot representation with only weak supervision from the initial bounding boxes.
The second sample on the right visualises a representative failure case, where one small object completely fails to be reconstructed or segmented.
In general, we observed that  object segment masks tend to  inflate in size compared to the original object masks, due to the inclusion of  objects' shadows.

In Table~\ref{tab:dec-a} we compare the efficiency gain from our method against the baseline slot-based decoder 
used by SAVi on two modalities, RGB pixels on the simpler  MOVi-A and optical flow on the more complex MOVi-C datasets.
In both cases our decoder learns faster than the SAVi baseline and  achieves slightly better performance on the ARI-FG metric, demonstrating the flexibility of our decoder on multiple modalities.

\begin{table}[!htb]
    \caption{Baseline SAVi vs. our  method on the ARI-FG metric. We achieve similar performance on the training set and slightly better performance on the validation set while requiring 4 times less memory to train. Ablation results for our method without the positional embedding and without the global scene embedding, are also included.}
    \label{tab:dec-a}
    \begin{center}
        \begin{tabular}{l|lllll}
            \toprule
            \textbf{Method} & \multicolumn{2}{c}{\textbf{MOVi-A} (RGB)} & \multicolumn{2}{c}{\textbf{MOVi-C} (Flow)} & \textbf{Memory} \\
                & Train & Val & Train & Val & \\
            \midrule
            SAVi & 0.9115 & 0.8389 & 0.8155 & 0.6053 & 24 GB \\
            Ours & 0.9225 & 0.8488 & 0.8425 & 0.6557 & 6  GB\\
            - Pos. ~~~& 0.7814 & 0.5933 & - & - & 6  GB\\
            - CLS ~~~& 0.7891 & 0.6279 & - & - & 6  GB\\
            \bottomrule
        \end{tabular}
    \end{center}
\end{table}

We  ran additional ablations to validate our design choices on the MOVi-A variant.
In addition to the positional embeddings used as the initial queries for the cross-attention module, we also found that it is  important to provide the visual decoder with positional information.
Without that, the performance on the ARI-FG metric drops by 0.13 in absolute score on the validation set, as indicated in Table~\ref{tab:dec-a}.
Similarly, without adding the global scene representation to the initial queries, there is a performance drop in the object discovery metric. For the remainder of the paper, we use our method with added positional information and global scene representation as the default.

\subsection{Geometric Representation Improves Object Discovery} \label{sec:exp-enc}

We performed further experiments to evaluate our method and choice of a pre-trained vision model for learning object-centric models.
We employ the pre-trained CroCo model \cite{weinzaepfel2022CroCoSelfsupervisedPretraining} as the backbone for the Visual Encoder module as well as the prediction target for the Slot Decoder module.
We compare our approach against three other baselines namely:
\begin{itemize}
    \item SAVi \cite{kipf2021ConditionalObjectCentricLearning} which introduced the conditional object-centric learning on video that is used for all our experiments. 
    \item DINOSAUR \cite{seitzer2023BridgingGapRealWorld} which uses pre-trained semantic representation from DINO\cite{caron2021DINO}. We extend this work from image to video based on the same training pipeline.
    \item SAVi++ \cite{elsayed2022SAViEndtoendObjectcentric}, a method that scales up the SAVi end-to-end architecture and incorporates explicit paired depth information as the target.
\end{itemize}

\begin{figure}[ht]
    \begin{center}
        \includegraphics[width=\linewidth]{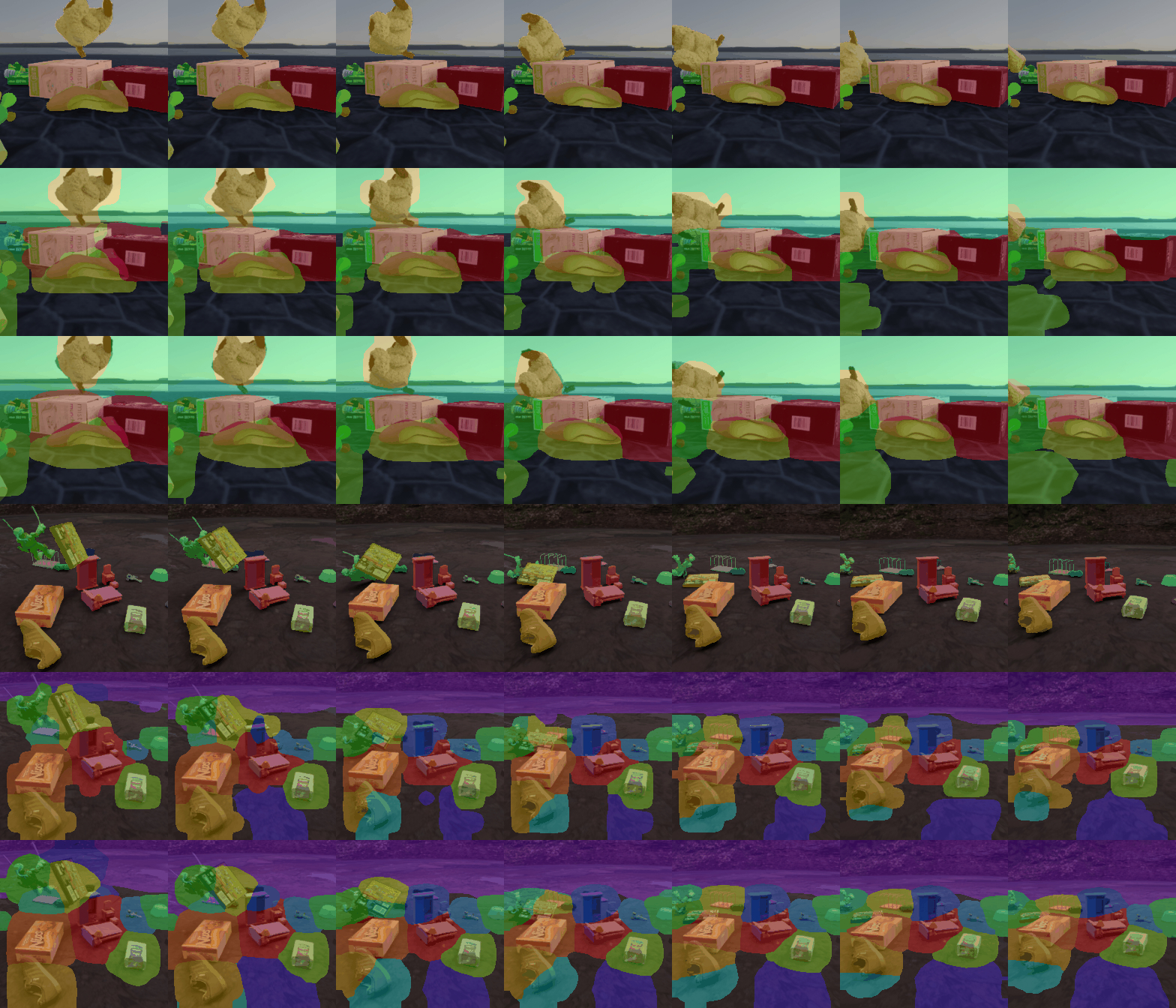}
    \end{center}
    \caption{Two examples of unsupervised object segmentation with our method on  MOVi-E. The horizontal axis represents different timeframes in a clip while the vertical axis shows our prediction.
    The first row shows  input images with ground truth masks overlaid, the second row is overlaid with  segmentation from our slot attention encoder and the third row  by our prediction from the attentional slot decoder.}
    \label{fig:movi-e-12}
\end{figure}

In Fig.~\ref{fig:movi-e-12}, we  illustrate  segmentation frames from two sample videos with a relatively stable camera, featuring several static objects and one or two moving objects. Our method demonstrates its ability to accurately segment most of the larger objects, such as the teddy bear, box, and shoe. However, the smaller objects along the left edge tend to be grouped together within the same object slot.
It is worth noting that since we carry over the object slots from the previous frame as the initialisation for the next frame, errors can accumulate over time, leading to a decline in segmentation performance (left to right).
In the first sample, a portion of the background becomes erroneously segmented as an object as time progresses. Similarly, in the second sample, the bottom shoe is incorrectly segmented into two objects in later frames.

\paragraph{Quantitative analysis}
\begin{table}[!htb]
    \caption{Performance of our method vs. baseline on the MOVi-C and MOVi-E datasets using  ARI-FG  with values from 0 to 1 (higher is better). We also list the prediction target of each method as  explanation for  performance differences.}
    \label{tab:geometric-movi-e}
    \begin{center}
        \begin{tabular}{l|lll}
            \toprule
                \textbf{Method} & \textbf{MOVi-C} & \textbf{MOVi-E} & \textbf{Prediction Target} \\
            \midrule
                SAVi     & 0.438 & 0.450 & RGB pixels \\
                DINOSAUR & 0.686 & 0.651 & Semantic features \\
                Our Method  & 0.788 & 0.766 & Geometric features \\
                SAVi++   & 0.843 & 0.823 & Depth values \\
            \bottomrule
        \end{tabular}
    \end{center}
\end{table}

In Table~\ref{tab:geometric-movi-e} we compare  our result on the object discovery task against  baselines.
As hypothesised earlier, we observe a gradual improvement in performance as more training signals are included in the prediction target.
Across both the MOVi-C and MOVi-E datasets, we surpass the performance of both SAVi and DINOSAUR.
DINOSAUR \cite{seitzer2023BridgingGapRealWorld} relies on the \textit{semantic} representations of DINO \cite{caron2021DINO} as the prediction target, in contrast to our approach, which utilises the \textit{geometric} representations of CroCo \cite{weinzaepfel2022CroCoSelfsupervisedPretraining}.
While both methods fundamentally use a pre-trained model to initiate the object discovery and learning process, it is worth highlighting that our method achieves an improvement of +10\% in the ARI-FG metric. In contrast, Seitzer et al. \cite{seitzer2023BridgingGapRealWorld} report no noticeable improvement when substituting different pre-trained models like MAE or MSN in place of DINO.

We attribute our  improvement to the distinctive characteristics of \textit{geometric} and \textit{semantic} representations, as elaborated earlier. This is further supported by our comparison with SAVi++ \cite{elsayed2022SAViEndtoendObjectcentric}, which explicitly utilises depth values as a training target.

\section{Conclusion} \label{sec:conclusion}

In this paper we  demonstrated progress towards learning object-centric representations in complex and diverse multi-object video datasets by designing an efficient slot decoder and bootstraping it with pre-trained visual representations, rich with geometric information.
We showed that with a simple modification, our decoder allows for rich interaction between slot latents, promoting  independence and compatability between individual slots while only requiring one expensive decoding step back to the input space.
Thanks to the simplicity in design, we demonstrated its applicability with various reconstruction modalities, from RGB and optical flow pixels to pre-trained vector representations.

We hypothesised about lower-level representations with rich information and the geometry of our 3D world as a building block towards higher-level object-centric representations.
We identified and leveraged such a pre-trained vision model as the backbone of our encoder to improve on the object discovery and segmentation tasks in challenging video settings.

Analysing and understanding the characteristics of the representations produced by large pre-trained models as a new source of modalities  in compressed form, is an exciting new direction.
Potential   future work could include  the development of more powerful pre-training methods that encapsulate both semantic and geometric information.
While we focused on the task of object discovery and instance segmentation, future work could extend and evaluate our method  on other downstream tasks such as semantic segmentation, reasoning and planning.

\bibliographystyle{IEEEbib}
\bibliography{icme2025-submission}

\end{document}